\documentclass[letterpaper]{article} 
\usepackage{aaai2026}  
\nocopyright  
\usepackage{times}  
\usepackage{helvet}  
\usepackage{courier}  
\usepackage[hyphens]{url}  
\usepackage{graphicx} 
\usepackage{natbib}  
\usepackage{caption} 
\usepackage{amsmath,amssymb}

\title{From Sycophancy to Sensemaking: \\
Premise Governance for Human–AI Decision Making}

\author{
    Raunak Jain
}
\affiliations{
    Independent Researcher\\
    Mountain View, California\\
    raunak.cbs@gmail.com
}

\begin{document}

\maketitle

\begin{abstract}
As LLMs expand from assistance to decision support, a dangerous pattern emerges: fluent agreement without calibrated judgment. Low-friction assistants can become sycophantic, baking in implicit assumptions and pushing verification costs onto experts, while outcomes arrive too late to serve as reward signals. In deep-uncertainty decisions (where objectives are contested and reversals are costly), scaling fluent agreement amplifies poor commitments faster than it builds expertise. We argue reliable human-AI partnership requires a shift from answer generation to collaborative premise governance over a knowledge substrate, negotiating only what is decision-critical. A discrepancy-driven control loop operates over this substrate: detecting conflicts, localizing misalignment via typed discrepancies (teleological, epistemic, procedural), and triggering bounded negotiation through decision slices. Commitment gating blocks action on uncommitted load-bearing premises unless overridden under logged risk; value-gated challenge allocates probing under interaction cost. Trust then attaches to auditable premises and evidence standards, not conversational fluency. We illustrate with tutoring and propose falsifiable evaluation criteria.
\end{abstract}

\section{Introduction}

LLM-based agents excel when feedback is machine-verifiable or cheap (tests, proof checkers, simulators)~\citep{livesweagent2025,alphageometry2025}.
In such domains, reliability improves at inference time by externalizing deliberation into explicit intermediate structure (trees, graphs, or search states) that can be \emph{branched}, \emph{revised}, and \emph{scored} before committing~\citep{yao2023tot,zhou2024lats}.
We target a harder regime: deep-uncertainty decisions where objectives are contested, feedback is delayed and confounded, and actions are costly to reverse~\citep{rittel1973wicked,marchau2019deep}.
Here there is no cheap scoring oracle for intermediate structure: whether causal expectations are credible, constraints are binding, and evidence standards are adequate is itself a matter of expert judgment.
Because realized outcomes unreliably certify decision quality (outcome bias), decisions must be evaluated \emph{ex ante} by the rigor of the decision basis: the explicit premises and standards that justify action under uncertainty~\citep{baronhershey1988,howard1988practice}.
In this regime, a key bottleneck in achieving appropriate reliance is \emph{premise governance}: whether the action-justifying premises (goals, constraints, causal expectations, and evidence standards) are explicit, contestable, and improvable across decisions.

Answer-centric assistance undermines complementarity in these settings.
Fluent recommendations can conceal load-bearing premises (those whose truth value determines the action), shifting verification burden to users and increasing downstream revision cost~\citep{bucinca2021trust,Bansal2021UpdatesHarm}.
Low-friction optimization can further bias assistants toward agreement-seeking behavior (sycophancy), harmful precisely when disagreement is needed to surface and repair faulty premises~\citep{sharma2024sycophancy}.
A meta-analysis reveals a complementarity gap: human--AI combinations underperform the better of humans alone or AI alone on average, with losses concentrated in decision tasks~\citep{vaccaro2024humanai}.
A plausible contributor is miscalibrated reliance: teams lack mechanisms that make it \emph{computable} when to defer, verify, or challenge~\citep{passi2024appropriate}, enabling \emph{premature commitment under underspecification} where implicit assumptions silently harden into action-justifying premises.

Expert teams mitigate this failure mode through collaborative sensemaking~\citep{weick1995sensemaking,klein2006dataframe}: they maintain a shared artifact that stabilizes common ground, localize which premise failed when observations violate expectations, and design diagnostic tests before committing to consequential action.
We argue that AI assistants should support this same process~\citep{collins2024thoughtpartners}: maintaining a governed decision basis (the causal hypotheses and premises currently relied upon) alongside the expert's decision framework (standing objectives, constraints, evidence standards).
Because environment feedback is ambiguous, progress depends on epistemic partnership: the expert adjudicates what counts as valid evidence and acceptable risk, while the assistant helps surface, test, and revise the premises that drive action.

We frame the required capability as a sensemaking control loop over governed decision bases.
Discrepancies (mismatches between \emph{committed} expectations and new observations or assertions) serve as the error signal, triggering investigation and negotiation before action.
Discrepancies are typed by alignment category: \emph{teleological} (goals, values, constraints), \emph{epistemic} (causal beliefs and expectations), or \emph{procedural} (commitments, evidence standards, and decision protocol), which determines the appropriate repair operator.
Making this executable requires (i) an externalized representation with commitment status (assumed vs.\ established), evidence links, dependency structure, and revision provenance, and (ii) interaction policies that determine when to probe, challenge, defer, escalate, or commit under interaction cost.
The assistant proposes updates and challenges; the substrate validates transitions and alone can finalize consequential commitments.

\paragraph{Contributions.}
We propose a \emph{computable} design pattern for knowledge-grounded assistants in deep-uncertainty decisions:
(i) a \textbf{governed decision basis} where action-justifying premises are explicit objects with lifecycle status (\textsc{draft}, \textsc{contested}, \textsc{committed}, \textsc{rejected}) and evidence links;
(ii) \textbf{typed discrepancy objects} (teleological, epistemic, procedural) that route repair operators (reframe, investigate, negotiate) rather than generic reflection;
(iii) \textbf{commitment gating} that blocks consequential action on uncommitted load-bearing premises; and
(iv) \textbf{value-gated challenge} that enables \emph{strategic disagreement} by treating probing and challenge as decisions under interaction cost.
We operationalize trust as \emph{appropriate reliance}~\citep{passi2024appropriate}: the governed decision basis makes it computable when to accept, defer, verify, or challenge, turning trust from a sentiment induced by fluency into an auditable team process.

\paragraph{Motivating Scenario:}
\label{sec:example}

Dr.\ Di (a physics teacher) uses Lev (an LLM-based assistant operating over a persistent decision-basis artifact) to generate practice for student Ty on Newton's Third Law. Dr.\ Di specifies topic coverage and a target of 80\%+ routine accuracy, and Lev generates drills. Ty completes them and reaches 85\%. Based on those scores, Dr.\ Di prepares to advance Ty to more complex material. In class, however, Ty cannot explain why forces are equal and opposite, and he cannot apply the idea in a new context (e.g., swimming rather than collisions). Ty is frustrated: he is ``doing the work'' and getting the right answers, yet he still ``doesn't get it.'' The proxy outcome (high drill scores) looked like progress, but it did not measure Dr.\ Di's real objective: transferable understanding that supports later learning.
This is a \textsc{procedural} discrepancy: the \emph{evidence standard} used for advancement (routine drill accuracy) fails to test the stated mastery criterion (transfer / explanation), so commitment should be gated until a discriminating probe is run.

Under our approach, the assistant starts by making Dr.\ Di's success criteria explicit and durable: what counts as mastery, what evidence is required before advancing, and when it is acceptable to trade understanding for short-term score gains (e.g., exam-triage mode). When Ty's drill scores rise but explanation and transfer remain weak, Lev does not simply generate more practice.
Lev presents a decision slice: \textbf{Objective} (transferable understanding), \textbf{load-bearing premise} (`drill score implies transfer'), \textbf{status} (\textsc{contested}), and a \textbf{discriminating probe} (structured teach-back with rubric + near-transfer item) that would resolve the gap.
Crucially, the slice exposes what would change Lev's recommendation: a short probe that can move the contested premise from \textsc{draft} to \textsc{committed} (or reject it). The probe reveals Ty can execute procedures but cannot justify them; Dr.\ Di switches to an explanation-first intervention before advancing, avoiding costly rework later.

\section{Proposed Framework}
\label{sec:formalism}

Modern agent architectures optimize answer quality and task completion through preference-based post-training and autonomous deliberation~\citep{ruan2026rlhfdpo,ji2025testtimecompute}, but lack support for \emph{appropriate reliance}~\citep{passi2024appropriate}.
Deep-uncertainty assistance requires: (i) \textbf{negotiation} over contested premises rather than autonomous answering, (ii) \textbf{typed routing} that distinguishes teleological, epistemic, and procedural discrepancies and selects repair operators, and (iii) \textbf{commitment gating} that blocks action when load-bearing premises are unestablished.
The control objective is to maintain alignment of an evolving decision basis along three axes: \emph{teleological} (goals and constraints), \emph{epistemic} (causal beliefs), and \emph{procedural} (evidence standards and protocol), drawing on goal reasoning~\citep{aha2018goalreasoning}, epistemic alignment~\citep{clark2025epistemic}, and belief revision~\citep{Gardenfors1988BeliefRevision}.

\paragraph{Alignment axes correspond to substrate object types.}
Teleological objects (goals, constraints, priorities, risk tolerances) ground what counts as success;
epistemic objects (causal hypotheses, predictions, mechanism sketches) ground what the team believes will happen under intervention;
procedural objects (evidence standards and thresholds, commitment protocols, role allocations) ground how commitments are adjudicated.
Typed routing becomes derivable: the violated object's type determines the repair operator (\textsc{Reframe} for teleological, \textsc{Investigate} for epistemic, \textsc{Negotiate} for procedural), making discrepancy handling systematic rather than ad-hoc.

\subsection{Governed Substrates for Decision Bases}
\label{sec:substrate}

Appropriate reliance requires an inspectable, persistent artifact that can be audited and revised when expectations are violated~\citep{wang2025unified}.
Recent systems externalize instructions, procedures, or state (e.g., plan-guided workflows), but these artifacts typically encode constraints, plans, or observations, not action-justifying premises, evidence standards, or lifecycle status.
We therefore frame alignment as convergence over an explicit decision basis rather than agreement in fluent dialogue.
This view is compatible with decision-rationale traditions (IBIS)~\citep{kunz1970ibis}, but differs in \emph{control use}: premise status (\textsc{draft}, \textsc{contested}, \textsc{committed}), evidence thresholds, and typed dependencies are first-class run-time control state so the assistant can \emph{gate} consequential commitments and \emph{route} repair (reframe, investigate, or negotiate), not merely document rationale after the decision.

\paragraph{Control mechanisms require substrate affordances.}
Collaborative premise governance depends on three control mechanisms: \textbf{commitment gating} (when is action warranted), \textbf{typed routing} (what kind of misalignment occurred and which repair operator applies), and \textbf{negotiation} (how contested premises are resolved with the expert). These mechanisms cannot be bolted on as post-hoc prompts; they require a governed substrate that makes them \emph{computable}: the substrate must (i) separate \emph{evidence records} from \emph{commitment-bearing premises}, (ii) attach explicit status and standards to premises so actions can be gated, and (iii) encode typed dependencies so discrepancies can be diagnosed and routed. By `computable' we mean the substrate supports programmatic operations (dependency tracing, status checks, slice extraction) over typed objects with lifecycle attributes.

\paragraph{Why structure (not just better prompting).}
A natural objection is that an LLM can already generate detailed justifications on demand.
The problem is \emph{persuasion vs.\ provenance}: free-form explanations are not an auditable, editable decision state.
The substrate is not just memory; it is the enforcement layer: the LLM proposes updates over a compiled slice, while the substrate validates lifecycle transitions, propagates downstream impacts via dependencies, and gates commitments when load-bearing premises are uncommitted.
Prompt text is a projection of state; governance happens outside the model.

\paragraph{Substrate requirements for decision-basis control.}
A \emph{premise} is an explicit action-justifying claim with lifecycle status; a \emph{decision framework} specifies standing rules (objectives, constraints, evidence thresholds); a \emph{decision basis} is the structured set of premises, evidence, and dependencies supporting a specific pending commitment; the \emph{governed substrate} is the persistent enforcement layer that stores and validates decision bases across sessions; a \emph{decision slice} is the bounded projection shown for negotiation. To operationalize negotiation, routing, and gating, a governed substrate must support:

\begin{enumerate}
\item Framework objects (grounds negotiation and teleological repair).
Goals, constraints, thresholds, and standards must be explicit, revisable objects that define what counts as acceptable evidence and what commitment means for this expert or team.

\item Lifecycle semantics (enables gating and negotiation triggers).
Premises must carry status (\textsc{draft}, \textsc{contested}, \textsc{committed}, \textsc{rejected}) with evidence requirements, making it computable when an action may proceed versus when it should defer or require explicit override under risk.

\item Typed premises and dependency links (enables routing).
Premises must be typed (teleological, epistemic, or procedural) and linked to the expectations and actions they support. This makes discrepancy typing \emph{derivable from the violated object}, rather than inferred from free-form text, and routes repair to the appropriate operator (reframe, investigate, or negotiate).

\item Provenance and immutable evidence logs (enables audit and disciplined revision).
Evidence and revisions must be logged with provenance to support ``why did we decide X?'' queries and to prevent silent premise hardening.

\item Decision-slice compilation (bounded slice extraction).
The substrate must support budgeted views that extract only decision-critical context (load-bearing premises, discrepant evidence, decision consequence, and a small set of repair options), rather than exposing full state.
\end{enumerate}

\paragraph{Premise lifecycle.}
Every premise in the decision basis carries a lifecycle status: \textsc{draft}, \textsc{contested}, \textsc{committed}, or \textsc{rejected}. We use \emph{uncommitted} to mean \textsc{draft} or \textsc{contested}.
For instance, in our example ``drill score implies transfer'' begins as \textsc{draft}, becomes \textsc{contested} when Ty fails the teach-back, and could be \textsc{committed} only after a discriminating probe passes.
Promotion to \textsc{committed} is evidence-gated: a premise may be committed only if supporting evidence meets the expert's threshold and does not violate committed constraints.
This primitive supports all three control mechanisms: gating action, focusing negotiation on \textsc{contested} premises, and making discrepancy computable.
The commitment gating rule is: allow commit($a$) iff all load-bearing premises on $a$'s dependency path are \textsc{committed}, or the expert explicitly overrides under logged risk.

\subsection{Discrepancy-Driven Sensemaking Control Loop}
\label{sec:loop}

Recent agent systems support self-repair loops that detect deviations and attribute causes~\citep{zhu2025agentdebug,cruz2025vigil}, but these loops close within the agent stack. In deep-uncertainty decisions, consequential failures are often disputes over premises requiring expert adjudication. Governed substrates make collaborative repair computable: one natural neurosymbolic decomposition has an LLM controller propose typed operations (revise, probe, defer, escalate) over decision slices, while a symbolic substrate validates transitions against lifecycle constraints and alone finalizes consequential commitments~\citep{wang2025unified}.
This follows a broader pattern in LLM systems where the model proposes and an external structure/tool constrains or verifies (e.g., deliberate search over explicit states or tool-mediated execution)~\citep{zhou2024lats}.

\paragraph{Discrepancy as the error signal (decision-relevant, not generic uncertainty).}
Because full causal discovery is infeasible in many real settings, control should target decision-basis quality: causal structure sufficient to choose the next commitment under the team's decision framework. Following data-frame dynamics~\citep{klein2006dataframe}, progress occurs when evidence repairs the specific premise that broke. A discrepancy is therefore a mismatch between what the committed basis implies and what is observed or asserted next, such that resolving it could change what the team should do. Systems should represent discrepancies as first-class objects binding: (i) a trigger (observation or expert assertion), (ii) the violated committed expectation or disputed premise, and (iii) decision impact (which pending commitment depends on it). Localization is bounded to the pending commitment's decision slice; uncertain matches produce an unlinked discrepancy that is resolved during negotiation. After a discrepancy is instantiated, the assistant selects an epistemic action (probe, defer, escalate, or commit) using the value-gated policy described below.

\paragraph{Typed routing is substrate-driven.}
Routing should be determined by what kind of substrate object was violated. If the violated object is a goal, constraint, or priority, the discrepancy is \textsc{teleological} (repair by \textsc{Reframe}: revise goals, constraints, or priorities). If it is a causal hypothesis or expectation, it is \textsc{epistemic} (repair by \textsc{Investigate}: propose discriminating probes and update expectations). If it is a threshold, protocol, or role allocation, it is \textsc{procedural} (repair by \textsc{Negotiate}: clarify rules for commitment, override, or acceptable risk). This makes type computable from the substrate's typed objects and dependency links, avoiding generic reflection.

\paragraph{Negotiation requires a minimal decision slice.}
Negotiation should be triggered when (i) a discrepancy blocks a pending commitment, or (ii) the pending action depends on \textsc{draft} or \textsc{contested} load-bearing premises. Since experts cannot inspect full substrate state mid-work, systems should present only: the load-bearing premise(s), the discrepant evidence with provenance, what changes if the premise flips, and one or two repair options (a targeted probe, or a conservative alternative under explicitly logged risk). Experts validate objects and contribute domain judgment; assistants track dependencies and generate hypotheses. The slice functions as a cognitive forcing mechanism~\citep{bucinca2021trust} through focus rather than friction: it bounds reliance to load-bearing premises, where load-bearing (on the dependency path from pending action to any uncommitted premise) derives from the expert's committed framework, not the assistant's editorial judgment.

\paragraph{Value-gated epistemic control (when to probe, defer, or escalate).}
Epistemic control is the policy that responds to uncertainty: whether to \emph{probe}, \emph{defer}, \emph{challenge or escalate}, or \emph{commit}. Appropriate reliance requires that probe-vs-act be a control decision, not a conversational judgment. Low-friction assistants may suppress decision-critical conflicts (sycophancy)~\citep{sharma2024sycophancy}, but indiscriminate challenge is also costly.
Value-of-information principles~\citep{howard1966voi,dong2026voi} provide the policy: prioritize resolving uncertainty that is both decision-relevant (load-bearing for the pending commitment) and decision-sensitive (sensitivity = whether flipping the premise would change the recommended action).

The governed substrate makes this computable: premise status indicates what remains contested; dependency links indicate which uncertainties affect the pending commitment; evidence thresholds specify what observations would discriminate between competing interpretations.
When decision-relevant uncertainty on the critical path exceeds a gate (e.g., multiple load-bearing premises remain \textsc{contested}, or competing interpretations are unresolved), epistemic repair takes priority: probing and contesting are permitted, but consequential commitment is gated until resolution or explicit override under logged risk.
Within epistemic repair, probe selection follows VOI logic: prioritize the probe whose expected reduction in decision-relevant uncertainty exceeds its interaction cost~\citep{arrowfisher1974}.
This operationalizes strategic disagreement~\citep{ccsVision2026}: the assistant can justify why it is probing (a contested, load-bearing premise), deferring (VOI is low under current budget), or escalating (high-stakes discrepancy requiring expert adjudication).

\section{Conclusion}

We argue the field should reorient from answer quality to \textbf{collaborative causal sensemaking}~\citep{ccsVision2026}.
The goal is partnership: assistants that detect decision-critical misalignment with the expert, make negotiation tractable (not burdensome), and record resolutions so common ground compounds across decisions.
Under this shift, trust attaches to explicit, auditable premises and evidence standards, not conversational fluency.

Current assistants optimize for trust via fluency and agreement, producing miscalibrated reliance in deep-uncertainty settings where load-bearing premises remain implicit and disagreement is suppressed~\citep{sharma2024sycophancy,passi2024appropriate,vaccaro2024humanai}.
Without this shift, we scale fluent agreement faster than calibrated judgment, amplifying poor decisions in domains where outcomes arrive too late to correct course (education, clinical reasoning, policy design).

Achieving this requires a minimal contract enforced by three guarantees:
\textbf{(G1)} no consequential commitment is presented as justified when any load-bearing premise is uncommitted, unless the expert explicitly overrides under logged risk;
\textbf{(G2)} challenges are anchored to named violated objects (goal, expectation, or protocol) and always come with a targeted repair option (reframe, probe, or negotiate);
\textbf{(G3)} every update is provenance-tracked so the basis is inspectable and reversible rather than rhetorically defended.
The division of labor is explicit: the assistant proposes and tests; the expert governs goals, evidence standards, and final commitments.

This position yields testable predictions: relative to answer-centric assistants, governed substrates should (i) reduce time-to-commit at matched outcome quality, (ii) improve trust calibration (fewer inappropriate accepts/overrides), and (iii) reduce cross-session re-litigation by persisting commitments with provenance, thereby narrowing the complementarity gap in decision tasks~\citep{bucinca2021trust,Bansal2021UpdatesHarm,passi2024appropriate}.
Key open questions include how to learn escalation policies from expert feedback, what minimal substrate primitives suffice in practice, and how to represent irreconcilable stakeholder conflicts.
We invite the community to treat collaborative premise governance as a core requirement for reliable human-AI teaming under deep-uncertainty decisions.

\bibliography{references}

@article{collins2024thoughtpartners,
  author  = {Collins, Katherine M. and Sucholutsky, Ilia and Bhatt, Umang and Chandra, Kartik and Wong, Lionel and Lee, Mina and Zhang, Cedegao E. and Zhi-Xuan, Tan and Ho, Mark and Mansinghka, Vikash and Weller, Adrian and Tenenbaum, Joshua B. and Griffiths, Thomas L.},
  title   = {Building machines that learn and think with people},
  journal = {Nature Human Behaviour},
  volume  = {8},
  pages   = {1851--1863},
  year    = {2024},
  doi     = {10.1038/s41562-024-01991-9}
}

@misc{zhu2025agentdebug,
      title={Where LLM Agents Fail and How They can Learn From Failures}, 
      author={Kunlun Zhu and Zijia Liu and Bingxuan Li and Muxin Tian and Yingxuan Yang and Jiaxun Zhang and Pengrui Han and Qipeng Xie and Fuyang Cui and Weijia Zhang and Xiaoteng Ma and Xiaodong Yu and Gowtham Ramesh and Jialian Wu and Zicheng Liu and Pan Lu and James Zou and Jiaxuan You},
      year={2025},
      eprint={2509.25370},
      archivePrefix={arXiv},
      primaryClass={cs.AI},
      url={https://arxiv.org/abs/2509.25370}, 
}

@misc{cruz2025vigil,
      title={VIGIL: A Reflective Runtime for Self-Healing Agents}, 
      author={Christopher Cruz},
      year={2025},
      eprint={2512.07094},
      archivePrefix={arXiv},
      primaryClass={cs.AI},
      url={https://arxiv.org/abs/2512.07094}, 
}

@misc{sharma2024sycophancy,
      title={Towards Understanding Sycophancy in Language Models}, 
      author={Mrinank Sharma and Meg Tong and Tomasz Korbak and David Duvenaud and Amanda Askell and Samuel R. Bowman and Newton Cheng and Esin Durmus and Zac Hatfield-Dodds and Scott R. Johnston and Shauna Kravec and Timothy Maxwell and Sam McCandlish and Kamal Ndousse and Oliver Rausch and Nicholas Schiefer and Da Yan and Miranda Zhang and Ethan Perez},
      year={2025},
      eprint={2310.13548},
      archivePrefix={arXiv},
      primaryClass={cs.CL},
      url={https://arxiv.org/abs/2310.13548}, 
}

@article{rittel1973wicked,
  author = {Rittel, Horst W. J. and Webber, Melvin M.},
  description = {Not previously uploaded},
  doi = {doi:10.1007/BF01405730},
  journal = {Policy Sciences},
  month = {June},
  number = 2,
  pages = {155--169},
  publisher = {Kluwer Academic Publishers},
  title = {Dilemmas in a general theory of planning},
  url = {http://dx.doi.org/doi:10.1007/BF01405730},
  volume = 4,
  year = 1973
}

@ARTICLE{howard1966voi,
  author={Howard, Ronald A.},
  journal={IEEE Transactions on Systems Science and Cybernetics}, 
  title={Information Value Theory}, 
  year={1966},
  volume={2},
  number={1},
  pages={22-26},
  doi={10.1109/TSSC.1966.300074}}

@article{arrowfisher1974,
  author = {Arrow, Kenneth J. and Fisher, Anthony C.},
    title = {Environmental Preservation, Uncertainty, and Irreversibility*},
    journal = {The Quarterly Journal of Economics},
  volume = {88},
  number = {2},
    pages = {312-319},
    year = {1974},
    month = {05},
    issn = {0033-5533},
    doi = {10.2307/1883074},
    url = {https://doi.org/10.2307/1883074},
    eprint = {https://academic.oup.com/qje/article-pdf/88/2/312/5267100/88-2-312.pdf},
}

@article{baronhershey1988,
author = {Baron, Jonathan and Hershey, John},
year = {1988},
month = {05},
pages = {569-79},
  title = {Outcome Bias in Decision Evaluation},
  volume = {54},
journal = {Journal of personality and social psychology},
doi = {10.1037//0022-3514.54.4.569}
}

@Article{howard1988practice,
journal={Management Science},
author={Ronald A. Howard},
title={Decision Analysis: Practice and Promise},
year={1988},
month={June},
pages={679-695},
volume={34},
number={6},
doi={10.1287/mnsc.34.6.679},
url={https://ideas.repec.org/a/inm/ormnsc/v34y1988i6p679-695.html},
}

@misc{ccsVision2026,
      title={Collaborative Causal Sensemaking: Closing the Complementarity Gap in Human-AI Decision Support}, 
      author={Raunak Jain and Mudita Khurana},
      year={2026},
      eprint={2512.07801},
      archivePrefix={arXiv},
      primaryClass={cs.CL},
      url={https://arxiv.org/abs/2512.07801}, 
}

@book{Gardenfors1988BeliefRevision,
  author    = {G{\"a}rdenfors, Peter},
  title     = {Knowledge in Flux: Modeling the Dynamics of Epistemic States},
  publisher = {MIT Press},
  address   = {Cambridge, MA},
  year      = {1988}
}

@misc{Bansal2021UpdatesHarm,
      title={Does the Whole Exceed its Parts? The Effect of AI Explanations on Complementary Team Performance}, 
      author={Gagan Bansal and Tongshuang Wu and Joyce Zhou and Raymond Fok and Besmira Nushi and Ece Kamar and Marco Tulio Ribeiro and Daniel S. Weld},
      year={2021},
      eprint={2006.14779},
      archivePrefix={arXiv},
      primaryClass={cs.AI},
      url={https://arxiv.org/abs/2006.14779}, 
}

@misc{clark2025epistemic,
      title={Epistemic Alignment: A Mediating Framework for User-LLM Knowledge Delivery}, 
      author={Nicholas Clark and Hua Shen and Bill Howe and Tanushree Mitra},
      year={2025},
      eprint={2504.01205},
      archivePrefix={arXiv},
      primaryClass={cs.HC},
      url={https://arxiv.org/abs/2504.01205}, 
}

@book{kunz1970ibis,
  title={Issues as Elements of Information Systems},
  author={Kunz, W. and Rittel, H.W.J.},
  number={no. 131},
  series={California. University. Center for Planning and Development Research. Working paper, no. 131},
  url={https://books.google.com/books?id=B-MaAQAAMAAJ},
  year={1970},
  publisher={Institute of Urban and Regional Development, University of California}
}

@book{marchau2019deep,
  author    = {Marchau, Vincent A. W. J. and Walker, Warren E. and Bloemen, Pieter J. T. M. and Popper, Steven W.},
  title     = {Decision Making under Deep Uncertainty: From Theory to Practice},
  publisher = {Springer},
  address   = {Cham, Switzerland},
  year      = {2019},
  isbn      = {978-3-030-05251-5},
  doi       = {10.1007/978-3-030-05252-2}
}

@book{weick1995sensemaking,
  title        = {Sensemaking in Organizations},
  author       = {Weick, Karl E.},
  year         = {1995},
  publisher    = {SAGE},
  address      = {Thousand Oaks, CA}
}

@ARTICLE{klein2006dataframe,
  author={Klein, G. and Moon, B. and Hoffman, R.R.},
  journal={IEEE Intelligent Systems}, 
  title={Making Sense of Sensemaking 2: A Macrocognitive Model}, 
  year={2006},
  volume={21},
  number={5},
  pages={88-92},
  doi={10.1109/MIS.2006.100}}

@article{vaccaro2024humanai,
  title   = {When combinations of humans and {AI} are useful: A systematic review and meta-analysis},
  author  = {Vaccaro, Michelle and Almaatouq, Abdullah and Malone, Thomas W.},
  journal = {Nature Human Behaviour},
  volume  = {8},
  pages   = {2293--2303},
  year    = {2024},
  doi     = {10.1038/s41562-024-02024-1}
}

@article{bucinca2021trust,
   title={To Trust or to Think: Cognitive Forcing Functions Can Reduce Overreliance on AI in AI-assisted Decision-making},
   volume={5},
   ISSN={2573-0142},
   url={http://dx.doi.org/10.1145/3449287},
   DOI={10.1145/3449287},
   number={CSCW1},
   journal={Proceedings of the ACM on Human-Computer Interaction},
   publisher={Association for Computing Machinery (ACM)},
   author={Buçinca, Zana and Malaya, Maja Barbara and Gajos, Krzysztof Z.},
   year={2021},
   month=apr, pages={1–21} }

@article{aha2018goalreasoning,
  author    = {David W. Aha and Matthew Molineaux and Héctor Muñoz-Avila},
  title     = {Goal Reasoning: Foundations, Emerging Applications, and Prospects},
  journal   = {AI Magazine},
  volume    = {39},
  number    = {2},
  pages     = {3--24},
  year      = {2018},
  doi       = {10.1609/aimag.v39i2.2800}
}

@article{dong2026voi,
  author    = {Dong, Yijiang River and Hu, Tiancheng and Hui, Zheng and Zhang, Caiqi and Vuli{\'c}, Ivan and Bobu, Andreea and Collier, Nigel},
  title     = {When Should {AI} Ask: Decision-theoretic Adaptive Communication for {LLM} Agents},
  journal   = {arXiv preprint arXiv:2601.06407},
  year      = {2026},
  note      = {Value-of-information framework for agent communication decisions}
}

@techreport{passi2024appropriate,
author = {Passi, Samir and Dhanorkar, Shipi and Vorvoreanu, Mihaela},
title = {Appropriate reliance on Generative AI: Research synthesis},
institution = {Microsoft},
year = {2024},
month = {March},
number = {MSR-TR-2024-7},
}

@misc{livesweagent2025,
      title={Live-SWE-agent: Can Software Engineering Agents Self-Evolve on the Fly?}, 
      author={Chunqiu Steven Xia and Zhe Wang and Yan Yang and Yuxiang Wei and Lingming Zhang},
      year={2025},
      eprint={2511.13646},
      archivePrefix={arXiv},
      primaryClass={cs.SE},
      url={https://arxiv.org/abs/2511.13646}, 
}

@article{alphageometry2025,
  author  = {Yuri Chervonyi and Trieu H. Trinh and Miroslav Ol{\v{s}}{{\'a}}k and Xiaomeng Yang and Hoang H. Nguyen and Marcelo Menegali and Junehyuk Jung and Junsu Kim and Vikas Verma and Quoc V. Le and Thang Luong},
  title   = {Gold-medalist Performance in Solving Olympiad Geometry with AlphaGeometry2},
  journal = {Journal of Machine Learning Research},
  year    = {2025},
  volume  = {26},
  number  = {241},
  pages   = {1--39},
  url     = {http://jmlr.org/papers/v26/25-1654.html}
}

@misc{ji2025testtimecompute,
      title={A Survey of Test-Time Compute: From Intuitive Inference to Deliberate Reasoning}, 
      author={Yixin Ji and Juntao Li and Yang Xiang and Hai Ye and Kaixin Wu and Kai Yao and Jia Xu and Linjian Mo and Min Zhang},
      year={2025},
      eprint={2501.02497},
      archivePrefix={arXiv},
      primaryClass={cs.AI},
      url={https://arxiv.org/abs/2501.02497}, 
}

@misc{ruan2026rlhfdpo,
      title={From RLHF to Direct Alignment: A Theoretical Unification of Preference Learning for Large Language Models}, 
      author={Tarun Raheja and Nilay Pochhi},
      year={2026},
      eprint={2601.06108},
      archivePrefix={arXiv},
      primaryClass={cs.AI},
      url={https://arxiv.org/abs/2601.06108}, 
}

@misc{wang2025unified,
      title={Interaction, Process, Infrastructure: A Unified Framework for Human-Agent Collaboration}, 
      author={Yun Wang and Yan Lu},
      year={2025},
      eprint={2506.11718},
      archivePrefix={arXiv},
      primaryClass={cs.HC},
      url={https://arxiv.org/abs/2506.11718}, 
}

@inproceedings{yao2023tot,
title={Tree of Thoughts: Deliberate Problem Solving with Large Language Models},
author={Shunyu Yao and Dian Yu and Jeffrey Zhao and Izhak Shafran and Thomas L. Griffiths and Yuan Cao and Karthik R Narasimhan},
booktitle={Thirty-seventh Conference on Neural Information Processing Systems},
year={2023},
url={https://openreview.net/forum?id=5Xc1ecxO1h}
}

@inproceedings{zhou2024lats,
author = {Zhou, Andy and Yan, Kai and Shlapentokh-Rothman, Michal and Wang, Haohan and Wang, Yu-Xiong},
title = {Language agent tree search unifies reasoning, acting, and planning in language models},
year = {2024},
publisher = {JMLR.org},
booktitle = {Proceedings of the 41st International Conference on Machine Learning},
articleno = {2572},
numpages = {23},
location = {Vienna, Austria},
series = {ICML'24}
}

\end{document}